\documentclass[journal,twoside,web]{ieeecolor}

\usepackage{jsen}
\usepackage{cite}
\usepackage{amsmath,amssymb,amsfonts}
\usepackage{algorithmic}
\usepackage{graphicx}
\usepackage{textcomp}
\usepackage{wrapfig}
\usepackage{makecell}
\usepackage{enumerate}
\usepackage{multirow}
\usepackage{bm}
\usepackage{textcomp}
\usepackage{footnote}
\usepackage{url}
\usepackage{pifont}
\usepackage{subfigure}
\usepackage{colortbl}

\newcommand{\cmark}{\ding{52}}%
\newcommand{\xmark}{\ding{56}}%
\newcommand{\sigla}[2]{#2~(#1)}%
\newcommand{\change}[1]{#1}
\definecolor{lightgreen}{rgb}{0.8, 1.0, 0.8}
\definecolor{lightred}{rgb}{1.0, 0.8, 0.8}

\def\BibTeX{{\rm B\kern-.05em{\sc i\kern-.025em b}\kern-.08em
    T\kern-.1667em\lower.7ex\hbox{E}\kern-.125emX}}
\definecolor{abstractbg}{rgb}{0.89804,0.94510,0.83137}
\begin{document}
\title{An Open-Source Tool for Classification Models in Resource-Constrained Hardware}
\author{Lucas Tsutsui da Silva, Vinicius M. A. Souza, and Gustavo E. A. P. A. Batista
\thanks{An earlier version of this paper was presented at the 2019 IEEE ICTAI Conference and was published in its proceedings: \protect\url{https://ieeexplore.ieee.org/document/8995408}.}
\thanks{This work was supported by the United States Agency for International Development (USAID), grant AID-OAA-F-16-00072, and the Brazilian National Council for Scientific and Technological Development (CNPq), grant 166919/2017-9.}
\thanks{Lucas Tsutsui da Silva is with Universidade de São Paulo, Brazil (e-mail: lucastsutsui@usp.br). }
\thanks{Vinicius M. A. Souza is with the Pontif\'{i}cia Universidade Cat\'{o}lica do Paran\'{a}, Brazil (e-mail: vinicius@ppgia.pucpr.br).}
\thanks{Gustavo E. A. P. A. Batista is with 
the University of New South Wales, Australia (e-mail: gbatista@cse.unsw.edu.au).}}

\IEEEtitleabstractindextext{%
\fcolorbox{abstractbg}{abstractbg}{%
\begin{minipage}{\textwidth}%
\begin{wrapfigure}[9]{r}{3.05in}%
\vspace{-0.12in}
\hspace{-0.10in}
\includegraphics[width=3.0in]{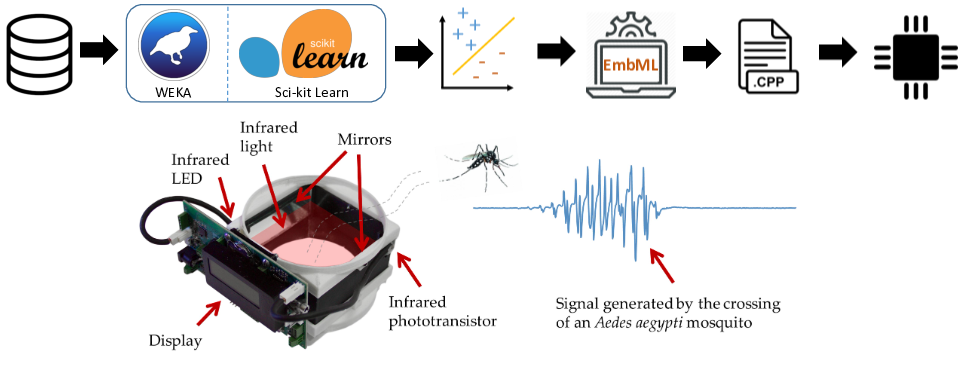}%
\end{wrapfigure}%
\begin{abstract}
Applications that need to sense, measure, and gather real-time information from the environment frequently face three main restrictions: power consumption, cost, and lack of infrastructure. Most of the challenges imposed by these limitations can be better addressed by embedding Machine Learning~(ML) classifiers in the hardware that senses the environment, creating smart sensors able to interpret the low-level data stream. However, for this approach to be cost-effective, we need highly efficient classifiers suitable to execute in unresourceful hardware, such as low-power microcontrollers. In this paper, we present an open-source tool named \textit{EmbML -- Embedded Machine Learning} that implements a pipeline to develop classifiers for resource-constrained hardware. We describe its implementation details and provide a comprehensive analysis of its classifiers considering accuracy, classification time, and memory usage. Moreover, we compare the performance of its classifiers with classifiers produced by related tools to demonstrate that our tool provides a diverse set of classification algorithms that are both compact and accurate. Finally, we validate EmbML classifiers in a practical application of a smart sensor and trap for disease vector mosquitoes.
\end{abstract}

\end{minipage}}}

\maketitle

\section{Introduction}
\label{sec:intro}

\IEEEPARstart{A}{pplications} that need to sense, measure, and gather real-time information from the environment frequently face three main restrictions~\cite{tubaishat2003sensor}: power consumption,  cost, and lack of infrastructure. For example, sensors often have a battery as their main power source, so efficient use of power allows them to run for extended periods. Price is a significant factor that hinders scaling in several areas, such as agriculture. Infrastructure assumptions, including \change{the} availability of reliable internet connection or power supply, frequently do not hold in a remote location or low-income countries.

Most of these challenges can be better addressed by embedding \sigla{ML}{Machine Learning} classifiers in the hardware that senses the environment, creating smart sensors able to interpret the low-level input. These smart sensors are low-powered systems that usually include one or more sensors, a processing unit, memory, a power supply, and a radio~\cite{yick2008wireless}. Since sensor devices have restricted memory capacity and can be deployed in difficult-to-access areas, they often use radio for wireless communication to transfer the data to a base station or to the cloud. Therefore, these smart sensors are more power-efficient since they eliminate the need for communicating all the raw data. Instead, they can only periodically report events of interest -- \textit{e.g.}, a dry soil crop area that needs watering or the capture of a disease-vector mosquito. 

However, for this approach to be cost-effective, we need highly efficient classifiers suitable to execute in sensors' unresourceful hardware, such as low-power microcontrollers. This scenario conflicts with the state-of-practice of ML, in which developers frequently implement classifiers in high-level interpreted languages such as Java or Python, make unrestricted use of floating-point operations and assume plenty of resources, including memory, processing, and energy. 

To overcome these problems, we present an open-source tool named \sigla{EmbML}{Embedded Machine Learning}~\cite{da2019embml} that implements a pipeline to develop classifiers for resource-constrained hardware. It starts with learning a classifier in a desktop or server computer using popular software packages or libraries such as \sigla{WEKA}{Waikato Environment for Knowledge Analysis}~\cite{hall2009weka} and scikit-learn~\cite{scikit-learn}. Next, EmbML converts the trained classifier into a carefully crafted C++ code with support for unresourceful hardware, such as the avoidance of unnecessary use of main memory and implementation of fixed-point operations for real numbers.

\change{EmbML} does not support the learning step in the embedded hardware. We advocate that, for most \change{ML} algorithms, the search for model parameters is too expensive to be performed on a microcontroller. However, most \change{ML} algorithms output classifiers that are highly efficient, including the ones supported in our tool: Logistic Regression, Decision Trees, Multilayer Perceptron~(MLP), and Support Vector Machine~(SVM).

\change{Our main contributions and findings are summarized below: 
\begin{itemize}
    \item We provide an open-source tool to convert ML models learned on a desktop/server using popular frameworks to their use on microcontrollers with constrained resources;
    \item We demonstrate the efficiency of EmbML in a public health case study and carry out a comprehensive experimental evaluation on six real-world benchmark datasets, four classes of ML algorithms, and six resource-constrained microcontrollers with varied characteristics;
    \item We empirically demonstrate that the use of fixed-point representation improves the classification time on microcontrollers without \sigla{FPU}{Floating-Point Unit} and can reduce the memory consumption of the classifiers;
    \item For neural network models, we show that the use of approximations for the sigmoid function can reduce its execution time when compared with the original function; 
    \item For tree-based models, we show that employing an if-then-else statement structure reduces the model execution time and does not impact memory consumption.
\end{itemize}
}

The remaining of this paper is organized as follows: Section~\ref{sec:rel-work} discusses related work; Section~\ref{sec:tool} presents the implementation details of EmbML; Section~\ref{sec:exp-setup} describes the experimental setup; in Section~\ref{sec:results} we assess the performance of EmbML; in Section~\ref{sec:modifications} we evaluate the modifications provided by EmbML; in Section~\ref{sec:compare-related-tools} we compare the EmbML performance; in Section~\ref{sec:intelligent-trap} we present a case study that employs our solution; \change{Section~\ref{sec:limitations} discusses the limitations of our work} and, finally, we present our conclusions in Section~\ref{sec:conclusion}.

\section{Related work}
\label{sec:rel-work}
EmbML has three main objectives: \textit{i}) having its source code available to the ML community for free usage and improvement; \textit{ii}) generating microcontroller-tailored classifier code with specific modifications to optimize its execution; and \textit{iii}) providing a variety of supported classification models.

\change{Various tools can convert ML models into source code for unresourceful hardware~\cite{sanchez2020tinyml}. However, most of them are from independent or industry developers, which leads to a scarcity of work with rigorous scientific analysis and experimental comparisons, as conducted in this paper. This section summarizes the most popular model conversion tools that can be directly compared to EmbML, allowing us to better establish our contribution to the state-of-the-art. As discussed, none of these works fulfill the three objectives of EmbML.}

Sklearn-porter\footnote{\url{https://github.com/nok/sklearn-porter}} is a popular tool to convert classification and regression models built off-board using the scikit-learn library. This tool supports a wide range of programming languages, including Java, JavaScript, C, Go, PHP, and Ruby, as well as several classifiers such as SVM, Decision Trees, Random Forest\change{~(RF)}, Naive Bayes\change{~(NB)}, \sigla{kNN}{k-Nearest Neighbors}, and MLP. Unfortunately, it does not provide any modification in the output classifier codes to support unresourceful hardware.

Weka-porter\footnote{\url{https://github.com/nok/weka-porter}} is a similar but more restricted project focused on the popular WEKA package. This tool converts J48 decision tree classifiers into C, Java, and JavaScript codes. Although the author indicates that this tool can be used to implement embedded classifiers, the lack of support for other classification algorithms restricts its applicability.

There are a considerable number of tools that transform decision tree models into C++ source code. The reason for such prevalence is the direct mapping of the models into if-then-else statements. Some examples are: J48toCPP\footnote{\url{https://github.com/mru00/J48toCPP}} that supports \textit{J48} classifiers from WEKA; C4.5 decision tree generator\footnote{\url{https://github.com/hatc/C4.5-decision-tree-cpp}} that converts \textit{C4.5} models from WEKA; and DecisionTreeToCpp\footnote{\url{https://github.com/papkov/DecisionTreeToCpp}} that converts \textit{DecisionTreeClassifier} models from scikit-learn.

SVM also has various conversion tools to C++. Two \change{examples} developed for microcontrollers are: mSVM\footnote{\url{https://github.com/chenguangshen/mSVM}} that includes support to fixed-point arithmetic;  and uLIBSVM\footnote{\url{https://github.com/PJayChen/uLIBSVM}} that provides a simplified version of \change{the SVM prediction function;} however, without support for fixed-point representation.

M2cgen\footnote{\url{https://github.com/BayesWitnesses/m2cgen}} converts ML models trained with scikit-learn into native code in Python, Java, C, JavaScript, PHP, R, Go, and others. It supports various  classification and regression models, including Logistic Regression, SVM, Decision Tree, \change{RF}, and XGBoost. Similarly to sklearn-porter, it does not provide any source code adaptation for microcontrollers.

Emlearn\footnote{\url{https://github.com/emlearn/emlearn}} generates code in C from decision tree, \change{NB}, MLP, and \change{RF} models built with scikit-learn or Keras. It includes features to support embedded devices, such as avoiding the usage of dynamic memory allocation and standard C library, as well as fixed-point representation for \change{NB} classifiers. It has little diversity of models, not supporting popular algorithms on embedded systems such as SVM. Also, the \change{NB} classifier is the only one currently able to use fixed-point arithmetic.

TensorFlow Lite for Microcontrollers\footnote{\url{https://www.tensorflow.org/lite/microcontrollers}} is a library designed to execute TensorFlow Neural Network~(NN) models on 32-bit hardware such as microcontrollers. To decrease the model size and memory usage, it allows applying post-training quantization, reducing the precision of numbers in the model. Some limitations identified by the authors include: support for a limited subset of TensorFlow operations, support for a limited set of devices, low-level C++ API requiring manual memory management, and lack of support for training models. 

EdgeML\footnote{\url{https://github.com/Microsoft/EdgeML/}} \change{enables generating} code from ML algorithms for resource-scarce devices. The library allows training, evaluation, and deployment of ML \change{models} on various devices and platforms. It \change{implements} modified and original ML algorithms that focus on time and memory efficiency to execute in resource-constrained devices~\cite{gupta2017protonn,dennis2018multiple,kusupati2018fastgrnn}. EdgeML also supports generating code that operates with fixed-point format~\cite{gopinath2019compiling}. Besides being a relatively complete solution, a drawback is \change{that it supports} ML models generated only by its original algorithms, \change{requiring} particular expertise to manipulate them. 

CMSIS-NN~\cite{lai2018cmsis} contains a set of efficient function implementations for layers usually \change{present} in NNs to help dump a trained NN in \change{ARM Cortex based microcontrollers}. Note that CMSIS-NN does not support training ML models, and the user is responsible for correctly combining function calls and uploading the network weights \change{into} the code. Also, this library implements \change{only} fixed-point operations and allows building an NN with any of the following layers: fully connected, convolution, pooling, softmax, and others. Disadvantages of using this library include supporting NN models \change{only}, and the manual process of producing a classifier code.

FANN-on-MCU~\cite{9016202} is a framework for the deployment of NNs trained with FANN library on ARM Cortex-M cores and parallel ultra-low power RISC-V-based processors. It offers automated code generation targeted to microcontrollers with fixed or floating-point formats and uses some implementations provided by CMSIS to improve performance on ARM Cortex-M cores. Though this is a robust solution, the number of supported ML models and microcontrollers is very restricted.

All the mentioned tools so far are open-source. An example of a proprietary tool is STM32Cube.AI which allows the conversion of NN models into optimized code that can run on STM32 ARM Cortex-M-based microcontrollers. It permits to import trained models from popular deep learning tools directly into STM32Cube.AI. Besides being a proprietary tool, another drawback is its lack of models diversity.

Commercial tools are abundant, but beyond the scope of this work due to the costs. Considering the popularity in academia, it is valid to mention the MATLAB Coder, a tool that converts a MATLAB program -- including ML models -- to produce C and C++ codes for a variety of hardware platforms.

Table~\ref{tab:related-tools} summarizes the related work. Although the description of each tool already includes some of \change{its} disadvantages compared to EmbML, the differences are more explicit when analyzing this table. For instance, EmbML is a solution that concomitantly offers: support to classification models from popular ML tools; diversity of efficient classifiers; and adaptations in the output code to improve its performance in resource-constrained hardware. \change{For a detailed survey, we redirect the reader to the work of Sanchez-Iborra and Skarmeta~\cite{sanchez2020tinyml}.}

\begin{table}[ht]
\setlength{\tabcolsep}{0.1cm}
\caption{Comparison between related tools.}
\label{tab:related-tools}
\centering
\resizebox{\columnwidth}{!}{%
\begin{tabular}{ccccc} \hline

\hline
\textbf{Tool} & {\textbf{\makecell{ML tool}}} & \textbf{Classifiers} & \textbf{Adaptations} &  \makecell{\textbf{Output code}} \\ \hline
EmbML & \makecell{WEKA and\\scikit-learn} & \makecell{Decision tree,\\SVM, MLP and\\logistic regression} & \makecell{fixed-point\\and sigmoid\\approximations} & C++ \\ 
sklearn-porter & scikit-learn & \makecell{SVM, kNN,\\decision tree,\\MLP and others} & -- & \makecell{Java, JavaScript,\\C, Go, PHP\\and Ruby} \\ 
weka-porter & WEKA & Decision tree & -- & \makecell{C, Java and\\JavaScript} \\ 
m2cgen & scikit-learn & \makecell{Logistic regression,\\decision tree,\\SVM and others} & -- & \makecell{Python, Java, C,\\JavaScript, Go,\\R and others} \\ 
emlearn & \makecell{scikit-learn\\and Keras} & \makecell{Decision tree, \change{NB},\\MLP and others} & \makecell{fixed-point\\for naive bayes} & C \\ 
\makecell{TensorFlow\\ Lite for\\ Microcontrollers} & TensorFlow & NN & quantization & C++ \\ 
EdgeML & -- & \makecell{Decision tree,\\kNN, and\\recurrent NN} & quantization & C and C++ \\ 
CMSIS-NN & -- & NN & fixed-point & C and C++ \\ 
FANN-on-MCU & FANN & NN & fixed-point & C \\ \hline

\hline
\end{tabular}%
}
\end{table}

\section{Proposed tool}
\label{sec:tool}

The main goal of EmbML\footnote{\url{https://github.com/lucastsutsui/EmbML}} is to produce classifiers source code suitable for executing in low-power microcontrollers. The process starts with creating a model using the WEKA package or the scikit-learn library from a dataset at hand. These are popular and open-source tools that provide a wide range of classification algorithms and simplify training and evaluating an ML model. After training the model in a desktop or server computer, the user needs to save it as an object serialized file capable of saving all the object content -- including the classifier parameters and data structures -- for future use.

EmbML receives such a serialized file as input and uses specific methods to deserialize the file. The deserialization process allows it to recover the classifier data and extract relevant parameters. Finally, EmbML fills a template file -- for a specific classifier -- using the data retrieved in the previous step and generates a C++ programming language file containing the model parameters, their initialization values, and implementations of the classification functions. This output file only includes the functions related to the classification step, since the user may later incorporate other functionalities according to the application, such as feature extraction and preprocessing, and further actions on the classifier output.

\begin{figure*}[htb]
    \centering
        \includegraphics[scale=0.4]{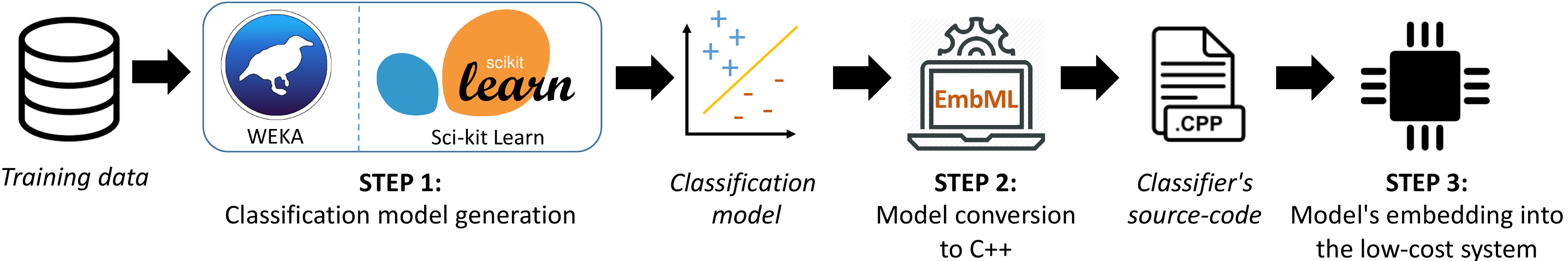}
    \vspace{-0.3cm}
    \caption{Workflow for generating classifier source codes using EmbML~\cite{da2019embml}.}
    \label{fig:tool}
\end{figure*}

Fig.~\ref{fig:tool} illustrates the operation workflow of EmbML. First, the user chooses one of the supported ML tools to process a training dataset and produce a classification model. EmbML is responsible for Step 2 in which it consumes the file containing the serialized model and creates the classifier source code. In this step, the user should decide to apply any of the provided modifications in the generated source code, such as using the fixed-point format to process real number operations. As we will further discuss, such a choice may impact both the accuracy and efficiency of the classifier. After evaluating the classifier in the desired hardware (Step 3), the user may return to Step 2 if they want to assess other modifications or to Step 1 if the classifier does not meet the application's requirements. In Step 3, it is possible to compile the code and deploy it on the microcontroller, for instance, using a combination of cross compilers and firmware upload protocols.

\subsection{Serialization and Model Recovery}

In the described pipeline, the user shall use serialization -- \textit{i.e.}, \change{to convert} an object state into a format so that the object can be stored in a file -- to provide \change{trained} classifiers as input to EmbML. Therefore, \change{serialization and deserialization allow to save a model trained with WEKA or scikit-learn} in a file and later recover it from this same file, respectively.

The WEKA classifier object must be serialized to a file using the \textit{ObjectOutputStream} and \textit{FileOutputStream} classes available in Java. To recover the model content, EmbML uses the \textit{javaobj}\footnote{\url{https://pypi.org/project/javaobj-py3/}} library -- available for the Python language. This library allows retrieving a Java serialized object from a file and producing a Python data structure, similar to the original object, that contains all its variables and data with unrestricted access, since Python has no private attributes. 

\change{For} scikit-learn models, it is possible to use the \textit{pickle}\footnote{\url{https://docs.python.org/3/library/pickle.html}} module that allows serializing and deserializing a Python object. \change{The user shall serialize the classification model into a file by applying the  \textit{dump} function.} After that, EmbML can recover the classifier object from this file using the \textit{load} method and access the object content without restriction.

\subsection{Algorithms and Classes}

The algorithms supported by EmbML are those suitable to execute in unresourceful hardware since they produce simple models that generally require little processing time and produce a small memory footprint. EmbML supports representative models of different learning paradigms: MLP networks, logistic regression, decision tree, and SVM.

EmbML accepts models from the following WEKA classes: \textit{J48} generates decision tree classifiers; \textit{Logistic} trains logistic regression classifiers; \textit{MultilayerPerceptron} produces MLP classifiers; and \textit{SMO} creates SVM classifiers -- with linear, polynomial, and \change{\sigla{RBF}{Radial Basis Function}} kernels.

It also supports the models from the following scikit-learn classes: \textit{DecisionTreeClassifier} produces decision tree models; \textit{LinearSVC} builds SVM classifiers with linear kernel; \textit{LogisticRegression} creates logistic regression classifiers; \textit{MLPClassifier} generates MLP classifiers; and \textit{SVC} trains SVM classifiers -- with polynomial and RBF kernels.

\subsection{General modifications}

To improve the classifiers' performance in low-power microcontrollers, EmbML implements modifications in all produced source codes. It is worth noting that EmbML never interferes with the training process, it only provides adjustments that affect the execution of the classifier source code. 

One modification is based on the idea that the model parameters -- \textit{e.g.}, the weights of an NN -- do not change during execution, according to our pipeline. Consequently, these data can be stored in the microcontroller's flash memory, once it is usually larger than its SRAM memory. \change{Therefore}, EmbML generates classifier source \change{codes that employ} the \textit{const} keyword, which expresses to the C++ compiler that these data are read-only and should be stored in the flash memory.

Another modification lies in the fact that most microcontrollers lack an FPU. This hardware is specifically designed to perform floating-point computations efficiently. When it is missing, processing real numbers becomes a challenging task. There are two \change{options} to tackle the absence of floating-point support~\cite{gopinath2019compiling}: emulating floating-point operations via software or converting real numbers to a fixed-point format. Software emulation usually is processing-expensive and results in a loss of efficiency. \change{However}, the second approach may reduce the range of representable values and cause a loss of precision.

EmbML enables producing classifier source codes that use both floating-point and fixed-point formats to store real numbers. For the first option, the generated code can directly proceed to the microcontroller's compilers since most of them already provide configuration options to emulate floating-point operations or use FPU instructions. For fixed-point representation, the scenario is quite different since the compilers of different microcontrollers do not offer a universal solution.

Therefore, we implemented a library of fixed-point operations based on: \textit{fixedptc}\footnote{\url{https://sourceforge.net/projects/fixedptc/}}, \textit{libfixmath}\footnote{\url{https://code.google.com/archive/p/libfixmath/}} and \textit{AVRfix}\footnote{\url{https://sourceforge.net/projects/avrfix/}}. It includes the basic arithmetic operations as well as other functions required by some classifiers -- \textit{e.g.}, exponential, power, and square root. Our library supports storing real numbers in integer variables with $32$, $16$, or $8$ bits and implements the \textit{Qn.m} format in which $n$ and $m$ are the numbers of bits in the integer and fractional parts, respectively~\cite{vluaductiu2012computer}.

\subsection{Modifications for MLP models}

EmbML supports \change{three different} approximations for the sigmoid function in MLP classifiers since it requires the expensive processing of the exponential function. The solution lies in replacing it, \change{in the inference step}, for functions that have similar behavior but perform simpler operations. \change{However, note that the model training still employs the original sigmoid function because scikit-learn and WEKA do not support defining custom activation functions.}

\begin{figure}[htb]
    \centering
        \includegraphics[scale=0.28]{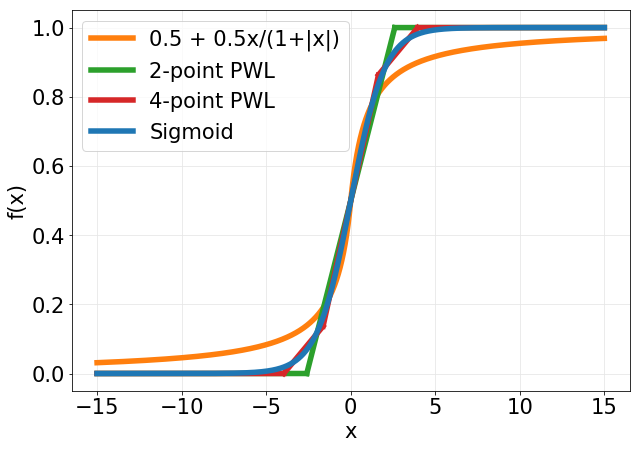}
    \vspace{-0.3cm}
    \caption{Approximations available in EmbML for the sigmoid function.}
    \label{fig:approx-sig}
\end{figure}

Fig.~\ref{fig:approx-sig} illustrates the three options alongside with the original sigmoid curve and reveals that they are in fact comparatively similar. Intuitively, the 4-point \sigla{PWL}{Piecewise Linear} version seems to be the most well-fitted version. The 2-point PWL produces a simple but relatively precise replica of the original pattern. The third approximation generates a smooth curve with strong correspondence with the sigmoid.

\change{Furthermore, to use processing and memory resources more efficiently, EmbML codes implement MLP models that reuse the output buffer of one layer as input to the next layer.}

\subsection{If-then-else statements for decision trees}

WEKA and scikit-learn source codes implement the classification steps of their decision tree models using iterative or recursive methods to traverse the tree structure. When EmbML processes a decision tree classifier, its default option is to produce an output code that contains the iterative version of tree traversal -- recursive methods are converted to their iterative form. EmbML provides an alternative representation for this process, which involves converting the binary tree structure into nested if-then-else statements. This approach intends to optimize the classifier time performance by eliminating the loop overhead of the iterative method -- \textit{e.g.}, instructions to increment the loop counter and test the end of the loop. \change{However}, it can consume more memory resources since its machine code will contain more instructions -- to reproduce the multiple comparisons from the if-then-else statements.

Finally, all the modifications employed by EmbML to optimize the \change{classifiers' processing times} and memory costs are summarized in Table~\ref{tab:code-modifications-embml}.

\begin{table}[ht]
\setlength{\tabcolsep}{0.1cm}
\caption{Code modifications supported by EmbML.}
\label{tab:code-modifications-embml}
\centering
\resizebox{\columnwidth}{!}{%
\begin{tabular}{ccccc}
\hline

\hline
\textbf{Classifier class} & \makecell{\textbf{\textit{Const}}\\ \textbf{variables}} & \makecell{\textbf{Fixed-point}\\ \textbf{representation}} & \makecell{\textbf{Sigmoid}\\ \textbf{approximations}} & \makecell{\textbf{If-then-else}\\ \textbf{statements}} \\ \hline
\textit{J48} & \cmark & \cmark & & \cmark \\
\textit{Logistic} & \cmark & \cmark & &  \\ 
\textit{MultilayerPerceptron} & \cmark & \cmark & \cmark & \\ 
\textit{SMO} & \cmark & \cmark & & \\ 
\textit{DecisionTreeClassifier} & \cmark & \cmark & & \cmark \\ 
\textit{LinearSVC} & \cmark & \cmark & & \\ 
\textit{LogisticRegression} & \cmark & \cmark & & \\
\textit{MLPClassifier} & \cmark &  \cmark & \cmark & \\ 
\textit{SVC} & \cmark & \cmark & & \\ \hline

\hline
\end{tabular}
}%
\end{table}

\section{Experimental setup}
\label{sec:exp-setup}

Our experiments consider three primary metrics to assess classifier performance: $i)$ accuracy rate, $ii)$ classification time, and $iii)$ memory usage. Ideally, we aim to obtain a high accuracy rate and low values for classification time and memory usage, since it may allow opting for simpler hardware -- reducing cost and power requirements for the system.

Accuracy and classification time are estimated using examples from a test set. Memory usage is measured from the compiled classifier code using the GNU \textit{size} utility. As for the classification time, we collected the mean value per instance using the \textit{micros} function from the Arduino library. To better estimate these values, we executed each classifier ten times in the test set. The microcontrollers always read the test set examples from a microSD memory card, but the results consider only the time spent in the classification process.

The analysis of EmbML classifiers includes the performance comparison of three representations for signed real numbers: 

\begin{enumerate}
    \item Floating-point with $32$ bits (referred as FLT), defined by the IEEE 754 standard and provided by the compiler;
    \item Fixed-point with $32$ bits (referred as FXP32), using the $Q22.10$ format provided by EmbML;
    \item Fixed-point with $16$ bits (referred as FXP16), using the $Q12.4$ format provided by EmbML.
\end{enumerate}

\subsection{Datasets}

We consider six benchmark datasets from real-world applications related to sensing, \change{representing} typical use-case scenarios for the EmbML classifiers. Follows a brief description of each dataset:

\begin{itemize}
\item Aedes aegypti-sex~\cite{dos2018unsupervised}. An optical sensor measures the wingbeat frequency and other \change{audio-related} features of flying insects that cross a sensor light. The classification task is to identify the sex of \textit{Aedes aegypti} mosquitoes;

\item Asfault-streets~\cite{souza2018asfault}. This dataset contains data collected from an accelerometer sensor installed in a car to evaluate pavement conditions of urban streets. The instances have the following categories related to the pavement quality: good, average, fair, and poor, as well as the occurrence of obstacles such as potholes or speed bumps;

\item Asfault-roads~\cite{souza2018asfault}. This dataset represents the same previously presented problem of pavement conditions evaluation but performed on highways instead of urban streets. The main \change{differences are} the lack of the poor class \change{and the} car speed during data collection;

\item GasSensorArray~\cite{vergara2012chemical}. \change{It} includes the data from a gas delivery platform with $16$ chemical sensors that measure six distinct pure gaseous substances in the air: ammonia, acetaldehyde, acetone, ethylene, ethanol, and toluene.

\item PenDigits~\cite{alimoglu1996methods}. It comprises the problem of classifying handwritten digits (from $0$ to $9$) according to the coordinates $(x, y)$ of a pen writing them on a digital screen;

\item HAR~\cite{anguita2013public}. This dataset contains data from accelerometer and gyroscope sensors of a waist-mounted smartphone for the following human activities: walking, climbing stairs, downstairs, sitting, standing, and lying down.

\end{itemize}

Table~\ref{tab:datasets} shows the datasets main characteristics. The experimental evaluation references these data using the identifiers.

\begin{table}[htb]
    \centering
    \setlength{\tabcolsep}{0.1cm}
    \caption{Characteristics of the evaluated datasets.}
    \label{tab:datasets}
    \begin{tabular}{llccc}
        \hline
        
        \hline
         \textbf{Identifier} & \textbf{Dataset} & \textbf{Features} & \textbf{Classes} & \textbf{Instances} \\ \hline
         D1 & Aedes aegypti-sex & $42$ & $2$ & $42,000$ \\
         D2 &Asfault-roads & $64$ & $4$ & $4,688$ \\         
         D3 &Asfault-streets  & $64$ & $5$ & $3,878$ \\
         D4 &GasSensorArray & $128$ & $6$ & $13,910$ \\
         D5 &PenDigits & $8$ & $10$ & $10,992$ \\
         D6 &HAR & $561$ & $6$ & $10,299$ \\ \hline
         
         \hline
    \end{tabular}
\end{table}

To evaluate the classifiers, we applied a $70/30$ holdout validation. This method splits the data into two stratified and mutually exclusive subsets: the training part takes $70\%$ of the instances, and the test set incorporates the $30\%$ remaining. 

\subsection{Classifiers}

We consider the default hyperparameter values provided by WEKA and scikit-learn. Therefore, the accuracy rates reported in our analysis may not represent the best possible values. \change{Also, the results obtained by the same model learned from different libraries cannot be directly compared since the hyperparameter values are not the same.} However, our primary concerns are to study the viability of embedded implementation of classifiers, determine if their execution in microcontrollers achieves the same accuracy as the desktop version, and optimize \change{classification} time and memory costs.

Given that we wanted to explore all possibilities supported by EmbML, there were a few cases in which we had to \change{set some hyperparameters manually}:

\begin{itemize}
    \item Since the \textit{SMO} class employs a linear kernel by default, we modified it to train the classifiers with polynomial (using $degree = 2$) and RBF kernels;
    \item In the case of \textit{SVC} class, we adjusted it to produce SVM models with a polynomial kernel (using $degree = 2$), since the RBF kernel is its default choice;
    \item We changed the \textit{MLPClassifier} \change{also to create} MLP networks that apply the sigmoid activation function, considering that the ReLU function is its default option.
\end{itemize}

\subsection{Microcontrollers}

Given the availability of a large number of microcontrollers suitable for low-power hardware, we selected six microcontrollers used in popular platforms for prototype projects. These platforms are representative examples that can be easily sourced, such as Arduino and Teensy, which also improves the reproducibility of the experiments. However, EmbML is not restricted to maker platforms by any means. Any microcontroller for which a C++ compiler is available can use the classifiers generated with the aid of EmbML.

The following microcontrollers were evaluated: 

\renewcommand{\theenumi}{\roman{enumi}}
\begin{enumerate}
    \item ATmega328/P available in the Arduino Uno. It is a low-power 8-bit microcontroller with the simplest features among the chosen models;
    \item ATmega2560 available in the Arduino Mega 2560. It is an 8-bit microcontroller \change{similar to the previous one, with some improvements in memory storage;}
    \item AT91SAM3X8E available in the Arduino Due. It is a high-performance 32-bit microcontroller and represents one of the most robust Arduino platforms;
    \item MK20DX256VLH7 available in the Teensy 3.1 and Teensy 3.2. It is a 32-bit microcontroller with intermediate processing and memory power;
    \item MK64FX512VMD12 available in the Teensy 3.5. It has a single-precision FPU, and \change{has better} clock frequency and memory storage compared to the previous version;
    \item MK66FX1M0VMD18 available in the Teensy 3.6. It is the most powerful processor in these experiments. \change{It} operates with 32 bits and includes a single-precision FPU.
\end{enumerate}

The Teensy platforms have an ARM Cortex-M4 core and the Arduino Due platform has an ARM Cortex-M3 core. The Arduino Uno and Arduino Mega 2560 have a low-power microcontroller from the AVR family. Table~\ref{tab:plataformas} shows some of the main specifications of the chosen embedded platforms.

\begin{table}[htb]
    \centering
    \setlength{\tabcolsep}{0.1cm}
    \caption{Characteristics of the evaluated embedded platforms.}
    \label{tab:plataformas}
    \resizebox{0.99\columnwidth}{!}{%
    \begin{tabular}{lccccc}
        \hline
        
        \hline
         \textbf{Platform} & \textbf{Microcontroller} & \makecell{\textbf{Clock}\\ \textbf{(MHz)}} & \makecell{\textbf{SRAM}\\ \textbf{(kB)}} & \makecell{\textbf{Flash}\\ \textbf{(kB)}} & \textbf{FPU} \\ \hline
         
         Arduino Uno & ATmega328/P & $20$ & $2$ & $32$ & \xmark \\
         Arduino Mega 2560 & ATmega2560 & $16$ & $8$ & $256$ & \xmark \\
         Arduino Due & AT91SAM3X8E & $84$ & $96$ & $512$ & \xmark \\
         Teensy 3.2 & MK20DX256VLH7 & $72$ & $64$ & $256$ & \xmark \\
         Teensy 3.5 & MK64FX512VMD12 & $120$ & $256$ & $512$ & \cmark \\
         Teensy 3.6 & MK66FX1M0VMD18 & $180$ & $256$ & $1,024$ & \cmark \\ \hline
         
        \hline
    \end{tabular}%
    }
\end{table}

In the following, we present our experimental evaluation considering three main aspects: accuracy, time, and memory; in Section~\ref{sec:modifications}, we assess the code modifications provided by EmbML; and Section~\ref{sec:compare-related-tools} presents a comparison of classifiers produced by EmbML and related tools. These sections contain an overview of the outcomes of our experiments. We provide detailed results in the supplementary material\footnote{\url{https://github.com/lucastsutsui/ieee-sensors-2021}} of this paper.

\section{Evaluation of EmbML classifiers}
\label{sec:results}

Next, we describe the experiments involving a sanity check and an analysis of classification time and memory consumption. First, we compare the accuracy rates obtained by EmbML classifiers running on the microcontrollers with the values obtained by scikit-learn or WEKA on a desktop computer, using the same test sets and corresponding trained models. Then, we estimate the time and memory results of the classifiers supported by EmbML as well as the impact of different representations for real numbers.

\subsection{Accuracy}

Table~\ref{tab:accuracy-classifiers} shows the accuracy rates for the test examples of each dataset running the models in a desktop and in a microcontroller with a classifier code produced by EmbML. It does not mention the microcontroller model since all results are the same, independent of the hardware. In these tables, the symbol ``-'' means that the produced code was too large and did not fit in any microcontroller's memory\change{, and the EmbML versions present relative differences to their desktop version.}

\begin{table}[htb]
    \centering
    \footnotesize
    \setlength{\extrarowheight}{-0.02cm}
    \setlength{\tabcolsep}{0.1cm}
    \caption{Accuracy ($\%$) for the EmbML classifiers.}
    \label{tab:accuracy-classifiers}
    \resizebox{0.99\columnwidth}{!}{%
    \begin{tabular}{cccccccc}
        \hline
        
        \hline
         \textbf{Classifier} & \textbf{Version} & \textbf{D1} & \textbf{D2} & \textbf{D3} & \textbf{D4} & \textbf{D5} & \textbf{D6} \\ \hline
        \multirow{4}{*}{\textit{J48}} 
        & Desktop & 99.00 & 88.48 & 84.28 & 97.41 & 84.71 & 94.34 \\ 
        & EmbML/FLT & 0.00 & 0.00 & 0.00 & 0.00 & 0.00 & 0.00  \\ 
        & EmbML/FXP32 & -0.03 & -0.07 & +0.26 & 0.00 & 0.00 & -0.33  \\ 
        & EmbML/FXP16 & -1.75 & -1.42 & -15.72 & -38.76 & 0.00 & -14.73 \\ \hline
        \multirow{4}{*}{\textit{Logistic}} 
        & Desktop & 97.71 & 91.61 & 89.00 & 98.97 & 73.00 & 97.35 \\ 
        & EmbML/FLT & 0.00 & 0.00 & 0.00 & 0.00 & 0.00 & 0.00  \\ 
        & EmbML/FXP32 & -0.06 & -0.07 & -1.03 & -0.62 & -0.28 & 0.00  \\ 
        & EmbML/FXP16 & -47.65 & -24.04 & -71.04 & -64.11 & -32.19 & -2.95 \\ \hline
        \multirow{4}{*}{\makecell{\textit{Multilayer-}\\ \textit{Perceptron}}} 
        & Desktop & 98.67 & 89.19 & 90.29 & 92.84 & 80.46 & 93.62 \\ 
        & EmbML/FLT & 0.00 & +0.07 & 0.00 & 0.00 & 0.00 & 0.00  \\ 
        & EmbML/FXP32 & -0.02 & +1.14 & +0.17 & +0.02 & +0.12 & +0.04  \\ 
        & EmbML/FXP16 & -44.27 & -0.57 & -1.80 & -74.46 & -0.58 & -0.90 \\ \hline
        \multirow{4}{*}{\makecell{\textit{SMO}\\(linear kernel)}} 
        & Desktop & 98.39 & 91.96 & 91.75 & 97.13 & 80.67 & 98.38 \\ 
        & EmbML/FLT & 0.00 & 0.00 & 0.00 & 0.00 & 0.00 & 0.00  \\ 
        & EmbML/FXP32 & +0.01 & +0.36 & +0.17 & 0.00 & -0.06 & +0.10  \\ 
        & EmbML/FXP16 & -8.42 & -10.52 & -20.36 & -74.78 & -2.33 & \cellcolor{lightred}\textbf{-81.65} \\ \hline
            \multirow{4}{*}{\makecell{\textit{SMO}\\(polynomial kernel)}} 
        & Desktop & 98.76 & 92.39 & 91.15 & 99.40 & 89.11 & 98.96 \\ 
        & EmbML/FLT & 0.00 & 0.00 & 0.00 & 0.00 & 0.00 & -  \\ 
        & EmbML/FXP32 & -0.05 & -1.35 & -1.63 & -62.46 & 0.00 & -  \\ 
        & EmbML/FXP16 & -39.73 & -65.08 & -51.37 & \cellcolor{lightred}\textbf{-85.36} & -44.63 & - \\ \hline
        \multirow{4}{*}{\makecell{\textit{SMO}\\(RBF kernel)}} 
        & Desktop & 98.08 & 87.62 & 83.59 & 75.59 & 67.63 & 95.99 \\ 
        & EmbML/FLT & 0.00 & 0.00 & - & - & - & - \\
        & EmbML/FXP32 & -0.09 & +0.15 & - & - & - & - \\
        & EmbML/FXP16 & -48.08 & -66.92 & -48.37 & - & -58.04 & - \\ \hline

        \multirow{4}{*}{\makecell{\textit{DecisionTree-}\\ \textit{Classifier}}}
        & Desktop & 98.54 & 86.13 & 84.02 & 97.03 & 83.83 & 93.20 \\
        & EmbML/FLT & -0.01 & 0.00 & 0.00 & 0.00 & 0.00 & 0.00  \\
        & EmbML/FXP32 & -0.05 & -0.35 & +0.26 & 0.00 & 0.00 & -0.35  \\ 
        & EmbML/FXP16 & -28.08 & -4.76 & -20.96 & -36.03 & 0.00 & -18.02 \\ \hline
        \multirow{4}{*}{\textit{LinearSVC}} 
        & Desktop & 90.51 & 92.11 & 88.83 & 80.02 & 36.74 & 98.58 \\ 
        & EmbML/FLT & 0.00 & 0.00 & 0.00 & 0.00 & 0.00 & 0.00  \\ 
        & EmbML/FXP32 & -3.87 & +0.07 & +0.09 & -44.75 & -0.33 & 0.00  \\ 
        & EmbML/FXP16 & -40.51 & \cellcolor{lightgreen}\textbf{-0.29} & -5.75 & -61.57 & -27.15 & -50.23 \\ \hline
        \multirow{4}{*}{\textit{LogisticRegression}} 
        & Desktop & 98.18 & 90.97 & 84.19 & 98.06 & 71.51 & 98.25 \\ 
        & EmbML/FLT & 0.00 & 0.00 & 0.00 & 0.00 & 0.00 & 0.00  \\ 
        & EmbML/FXP32 & -0.03 & -0.07 & -0.08 & -51.89 & +0.24 & +0.03  \\ 
        & EmbML/FXP16 & -48.18 & \cellcolor{lightgreen}\textbf{-0.07} & -0.77 & -79.61 & -31.13 & \cellcolor{lightgreen}\textbf{-0.13} \\ \hline
        \multirow{4}{*}{\textit{MLPClassifier}} 
        & Desktop & 95.96 & 92.46 & 91.41 & 96.43 & 89.96 & 98.54 \\ 
        & EmbML/FLT & 0.00 & 0.00 & 0.00 & 0.00 & 0.00 & 0.00  \\ 
        & EmbML/FXP32 & +0.16 & +0.14 & +0.43 & -0.17 & -0.09 & -0.16  \\ 
        & EmbML/FXP16 & -39.52 & \cellcolor{lightred}\textbf{-87.34} & -27.32 & -79.76 & -32.19 & -60.22\\ \hline
        \multirow{4}{*}{\makecell{\textit{SVC}\\(polynomial kernel)}} 
        & Desktop & 98.47 & 77.17 & 64.78 & 98.87 & 90.75 & 93.95 \\ 
        & EmbML/FLT & -46.91 & 0.00 & 0.00 & -1.84 & -19.24 & -  \\ 
        & EmbML/FXP32 & -48.25 & 0.00 & 0.00 & -80.42 & -81.56 & - \\ 
        & EmbML/FXP16 & -48.47 & -0.36 & -29.56 & -80.42 & -80.62 & - \\ \hline
        \multirow{4}{*}{\makecell{\textit{SVC}\\(RBF kernel)}} 
        & Desktop & 58.53 & 88.62 & 86.51 & 21.63 & 18.69 & 95.28 \\ 
        & EmbML/FLT & - & 0.00 & 0.00 & - & 0.00 & - \\ 
        & EmbML/FXP32 & - & +0.14 & -0.43 & - & -0.36 & - \\ 
        & EmbML/FXP16 & -8.53 & -67.92 & -51.29 & 0.00 & -8.80 & -76.41 \\ \hline
        
        \hline
    \end{tabular}%
}
\end{table}

As expected, the classifiers using FLT obtain the same accuracy rates than their desktop counterparts. These results imply that the EmbML classifiers correctly implement the trained models. There are only some minor exceptions:

\begin{itemize}
    \item For D2 with the \textit{MultilayerPerceptron}, the accuracy increases from $89.19\%$ (in desktop) to $89.26\%$ (with EmbML classifier using FLT);
    \item For D1 with the \textit{DecisionTreeClassifier}, the accuracy reduces from $98.54\%$ (in desktop) to $98.53\%$ (with EmbML classifier using FLT);
    \item For D1, D4, and D5 with \textit{SVC} (polynomial kernel), the accuracies from EmbML classifiers (using FLT) are lower than those obtained in desktop. In this specific case, the decrease happens because the \textit{SVC} employs double-precision (64 bits) floating-point operations, and EmbML only supports single-precision (32 bits). \change{We confirmed this fact by executing, on a desktop, the EmbML output codes for these cases using both representations.}
\end{itemize}

\change{The reduction in precision for \textit{SVC} models affects intermediate calculations depending on various factors intrinsic to each dataset (\textit{e.g.}, the ranges of attribute values and model weights) that also explain unstable results for a given model on different datasets.} \change{Moreover}, in most cases, there is not a significant change in accuracy when using FXP32 \change{compared} to FLT. On the other hand, \change{the use of} FXP16 can cause a notable reduction in accuracy for most classifiers due to several reasons. \change{For instance, underflow\footnote{\change{We consider underflow when it rounds a non-zero real number to zero, possibly canceling out results of subsequent multiplications.}} and overflow occur regularly (from $26.64\%$ to $38.71\%$) in the arithmetic operations of the cases highlighted in red in Table~\ref{tab:accuracy-classifiers}, associated with high accuracy losses. Differently, they happen less frequently (from $14.78\%$ to $19.07\%$) in the operations of the cases highlighted in green, associated with low accuracy losses.} \change{Also, the results imply that FXP16 may be more sensitive to the choices of $n$ and $m$, which are not optimal in our experiments and can negatively affect accuracy.}

\subsection{Classification Time}

\begin{figure}[htb]
    \centering
        \includegraphics[width=0.97\columnwidth]{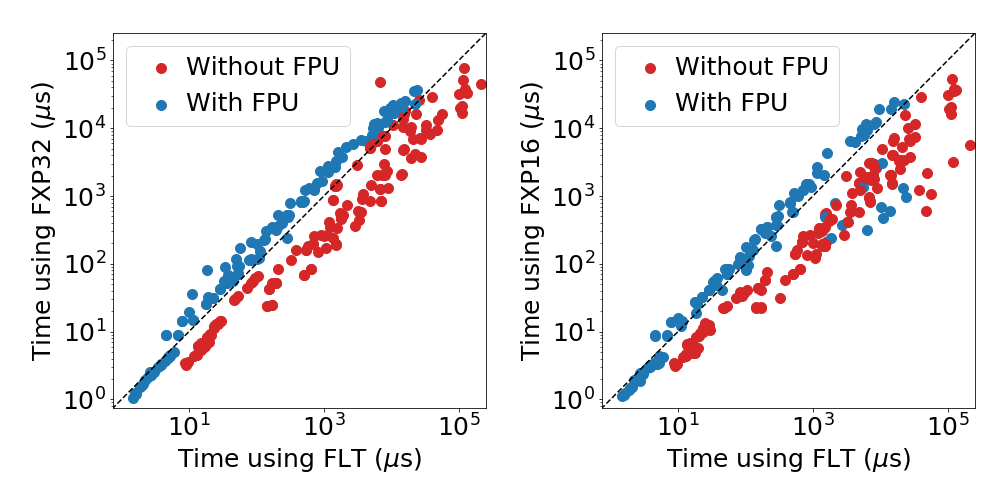}
    \vspace{-0.4cm}
    \caption{Run-time comparison for floating-point and fixed-point formats for FXP32 (left) and FXP16 (right).}
    \label{fig:time-flt-fxp}
\end{figure}

Next, we compare the average time that each model spent to classify an instance. Fig.~\ref{fig:time-flt-fxp} shows the executions of the EmbML codes using different representations for real numbers. In these graphs, the coordinates of each point represent the results of average classification time of FLT and FXP32 (left); and FLT and FXP16 (right) -- for the same classifier, microcontroller, and dataset. In the microcontrollers that lack an FPU, we can observe that fixed-point versions achieve lower classification time than FLT. \change{However, we do not see such an improvement in microcontrollers with FPU.} 

\begin{figure}[htb]
    \centering
        \includegraphics[scale=0.28]{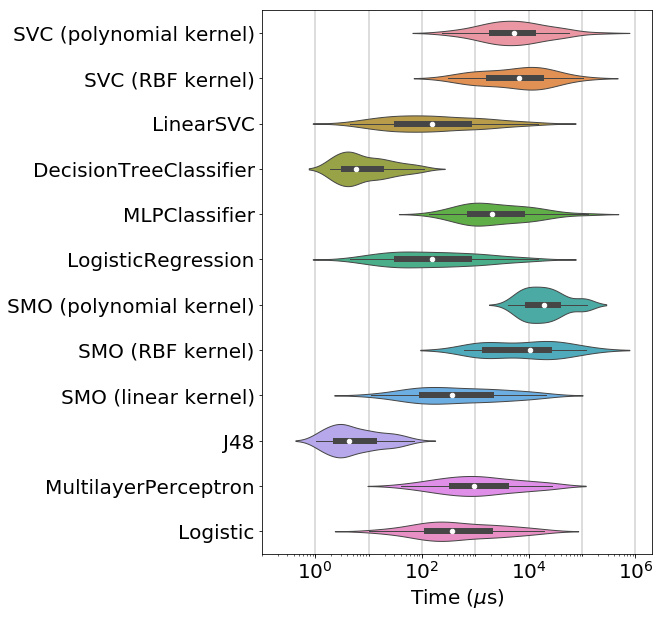}
    \vspace{-0.2cm}
    \caption{Run-time comparison for all classifiers.}
    \label{fig:time-results}
\end{figure}

In Fig.~\ref{fig:time-results}, we present the average classification time per classifier class. This graph combines all results of each classifier obtained from executing it with every microcontroller and dataset and presents them in a way to contrast their behavior. From it, we can notice that the decision tree models usually deliver the lowest classification time. The logistic regression and SVM (linear kernel) models have similar performances, reaching the overall second-best results. The MLP models achieve an intermediate performance: faster than the SVM with polynomial and RBF kernels, but usually slower than the others. The SVM models with the polynomial and RBF kernels perform the worst results.

\subsection{Memory Usage}

\begin{figure}[htb]
    \centering
        \includegraphics[width=0.97\columnwidth]{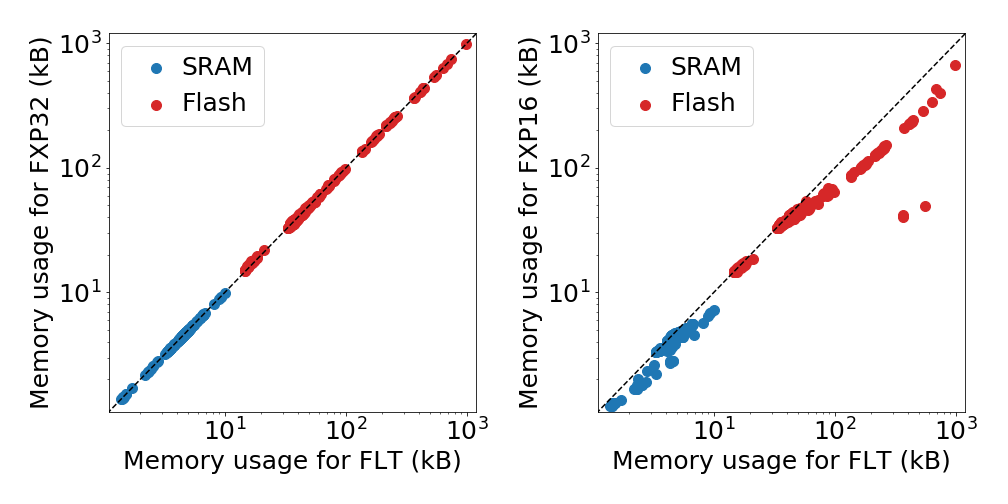}
    \vspace{-0.4cm} 
    \caption{Memory usage comparison for floating-point and fixed-point formats for FXP32 (left) and FXP16 (right).}
    \label{fig:mem-flt-fxp}
\end{figure}

The last metric evaluated from classifier performance is memory consumption. We \change{separately} compare data (SRAM) and program (flash) memories among the supported classifiers. In Fig.~\ref{fig:mem-flt-fxp}, we show the memory usage of classifiers using FXP32 (left) and FXP16 (right) compared to FLT. These graphs reveal that there is no advantage of employing FXP32 in terms of this analysis. However, FXP16 representation can decrease memory consumption.

\begin{figure}[htb]
    \centering
        \includegraphics[scale=0.28]{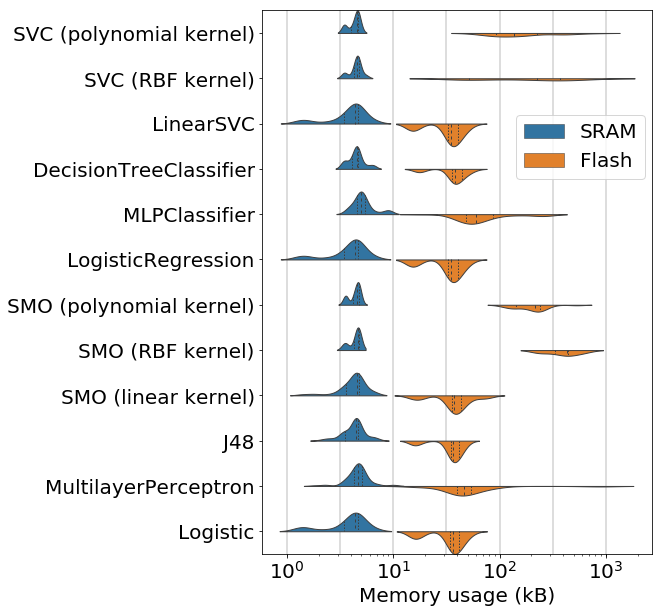}
    \vspace{-0.2cm}
    \caption{Memory usage comparison for all classifier.}
    \label{fig:mem-results}
\end{figure}

Fig.~\ref{fig:mem-results} displays the analysis of memory usage per classifier class, including results from all combinations of microcontrollers and datasets. \change{We can observe that} the decision tree, logistic regression, and SVM (linear kernel) models achieve the best memory performance. On the other hand, the SVM classifiers with polynomial and RBF kernels obtain the highest memory consumption. \change{The MLP models} again have an intermediate performance in \change{most cases compared} to the two previously defined groups.

\section{Evaluation of code modifications}
\label{sec:modifications}

To understand if the proposed modifications positively affect the time performance and do not cause a negative effect on the accuracy rates, we experimentally evaluate the impact of each one. It includes the analysis of approximations for sigmoid function in MLP models and \change{transforming decision trees} from an iterative structure to if-then-else statements. Since the previous section already assessed the impact of using fixed-point, this modification is not individually analyzed.

\subsection{Approximations for Sigmoid Function in MLP}

Table~\ref{tab:accuracy-mlp-opt-weka} and Table~\ref{tab:accuracy-mlp-opt-sklearn} present the accuracy rates estimated in each test set through applying the approximation functions provided by EmbML for substituting the sigmoid in MLP models of WEKA and scikit-learn, respectively. These tables contain the accuracy values for the different real numbers representations and also the accuracy obtained by MLP models using the original sigmoid function for comparison. \change{The values from EmbML versions are shown as the relative difference to the desktop version with the original sigmoid.}

\begin{table}[htb]
    \centering
    \footnotesize
    \setlength{\extrarowheight}{-0.02cm}
    \setlength{\tabcolsep}{0.1cm}
    \caption{Accuracy ($\%$) for the \textit{MultilayerPerceptron} models.}
    \label{tab:accuracy-mlp-opt-weka}
    \resizebox{0.99\columnwidth}{!}{%
    \begin{tabular}{cccccccc}
        \hline
    
        \hline
         \textbf{Classifier} & \textbf{Version} & \textbf{D1} & \textbf{D2} & \textbf{D3} & \textbf{D4} & \textbf{D5} & \textbf{D6} \\ \hline
        \multirow{4}{*}{\makecell{Original\\sigmoid}}
        & Desktop & 98.67 & 89.19 & 90.29 & 92.84 & 80.46 & 93.62 \\ 
        & EmbML/FLT & 0.00 & +0.07 & 0.00 & 0.00 & 0.00 & 0.00  \\ 
        & EmbML/FXP32 & -0.02 & +1.14 & +0.17 & +0.02 & +0.12 & +0.04  \\ 
        & EmbML/FXP16 & -44.27 & -0.57 & -1.80 & -74.46 & -0.58 & -0.90 \\ \hline
        \multirow{3}{*}{\makecell{$0.5 + 0.5x/(1+|x|)$\\function}}
        & EmbML/FLT & 0.00 & 0.00 & +0.09 & +0.07 & 0.00 & +0.07  \\ 
        & EmbML/FXP32 & -0.02 & 0.00 & +0.17 & +0.14 & +0.03 & +0.10  \\ 
        & EmbML/FXP16 & -44.32 & -1.92 & -1.80 & -74.18 & -0.94 & -0.19 \\ \hline
        \multirow{3}{*}{2-point PWL} 
        & EmbML/FLT & 0.00 & +1.71 & -0.08 & -0.12 & -0.27 & +0.07  \\ 
        & EmbML/FXP32 & -0.02 & +1.85 & -0.17 & -0.08 & -0.24 & +0.07  \\ 
        & EmbML/FXP16 & -44.28 & -0.50 & -2.15 & -74.32 & -0.48 & -0.93 \\ \hline
        \multirow{3}{*}{4-point PWL}
        & EmbML/FLT & 0.00 & +1.78 & +0.26 & +0.02 & -0.06 & +0.07  \\ 
        & EmbML/FXP32 & -0.02 & +1.78 & +0.09 & +0.02 & -0.09 & +0.04  \\ 
        & EmbML/FXP16 & -44.28 & -0.78 & -1.63 & -74.56 & -0.30 & -0.93 \\ \hline
        
        \hline
    \end{tabular}%
}
\end{table}

\begin{table}[htb]
    \centering
    \footnotesize
    \setlength{\extrarowheight}{-0.02cm}
    \setlength{\tabcolsep}{0.1cm}
    \caption{Accuracy ($\%$) for the \textit{MLPClassifier} models with sigmoid.}
    \label{tab:accuracy-mlp-opt-sklearn}
    \resizebox{0.99\columnwidth}{!}{%
    \begin{tabular}{cccccccc}
        \hline
        
        \hline
         \textbf{Classifier} & \textbf{Version} & \textbf{D1} & \textbf{D2} & \textbf{D3} & \textbf{D4} & \textbf{D5} & \textbf{D6} \\ \hline
        \multirow{4}{*}{\makecell{Original\\sigmoid}} 
        & Desktop & 98.39 & 92.25 & 90.72 & 74.58 & 91.78 & 98.64 \\ 
        & EmbML/FLT & 0.00 & 0.00 & 0.00 & 0.00 & 0.00 & 0.00  \\ 
        & EmbML/FXP32 & -0.17 & -0.07 & 0.00 & -0.19 & +0.06 & +0.07  \\ 
        & EmbML/FXP16 & -17.18 & -2.42 & -10.99 & -56.97 & -3.98 & -57.83 \\ \hline
        \multirow{3}{*}{\makecell{$0.5 + 0.5x/(1+|x|)$\\function}}
        & EmbML/FLT & 0.00 & -0.93 & -0.43 & -0.07 & -0.18 & +0.16  \\ 
        & EmbML/FXP32 & -0.13 & -0.93 & -0.51 & -0.19 & -0.24 & +0.07  \\ 
        & EmbML/FXP16 & -11.57 & -2.07 & -8.16 & -57.26 & -5.28 & -39.55 \\ \hline
        \multirow{3}{*}{2-point PWL}
        & EmbML/FLT & -0.03 & +0.28 & -0.34 & 0.00 & -0.03 & -0.06  \\ 
        & EmbML/FXP32 & -0.21 & +0.21 & -0.08 & -0.19 & +0.09 & -0.06  \\ 
        & EmbML/FXP16 & -16.50 & -2.63 & -9.71 & -57.02 & -3.86 & -59.51 \\ \hline
        \multirow{3}{*}{4-point PWL}
        & EmbML/FLT & +0.01 & -0.22 & +0.26 & 0.00 & +0.03 & -0.06  \\ 
        & EmbML/FXP32 & -0.20 & -0.14 & +0.52 & -0.19 & +0.09 & 0.00  \\ 
        & EmbML/FXP16 & -16.78 & -4.13 & -14.09 & -56.97 & -3.58 & -56.92 \\ \hline
        
        \hline
    \end{tabular}%
}
\end{table}

\change{These tables must be analyzed independently because WEKA and scikit-learn use different hyperparameter values to train a model.} Contrasting the alternative modifications with the original sigmoid, the highest difference in accuracy for the \textit{MultilayerPerceptron} models occurred in D2 using the 4-point PWL approximation and FLT. In this case, the accuracy rate increased from $89.26 \%$ \change{(original sigmoid)} to $90.97 \%$ \change{(4-point PWL)}. As for the \textit{MLPClassifier}, the maximum difference happened in D6 using the $0.5 + \frac{0.5x}{(1+|x|)}$ function and FXP16. The accuracy increased from $40.81 \%$ \change{(original sigmoid)} to $59.09 \%$ \change{(approximation function)} in this situation. As a general rule, the accuracy values from the modified models are relatively close to the original versions and can be acceptable in practice.

\begin{figure}[htb]
    \centering
        \includegraphics[width=0.45\columnwidth]{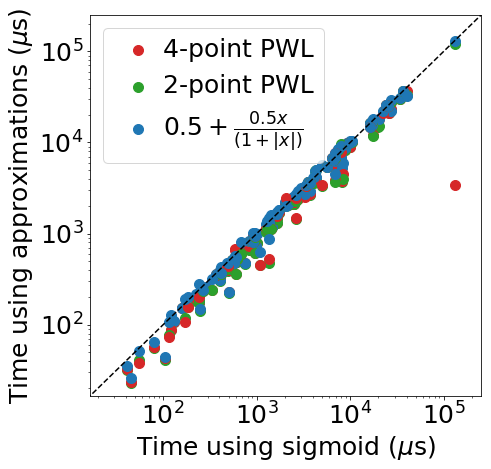}
    \vspace{-0.2cm}
    \caption{Time comparison for approximation functions in MLP models.}
    \label{fig:time-sig-approx}
\end{figure}

In Fig.~\ref{fig:time-sig-approx}, we exhibit the average classification time comparison for MLP models using the provided options for the sigmoid function. We can identify that these options produce similar time results in many cases. However, the use of PWL approximations can frequently decrease the classification time of MLP models, whereas not causing expressive changes in the accuracy rates. Consequently, these versions are attractive options to help improve the performance of MLP classifiers.

We do not include memory usage analysis, since the difference in this metric for using or not the modified classifiers is relatively inexpressive, considering that the sigmoid approximations do not affect the size of the classifier variables.

\subsection{If-Then-Else Statements and Iterative Decision Trees}

\begin{figure}[htb]
    \centering
        \includegraphics[width=0.45\columnwidth]{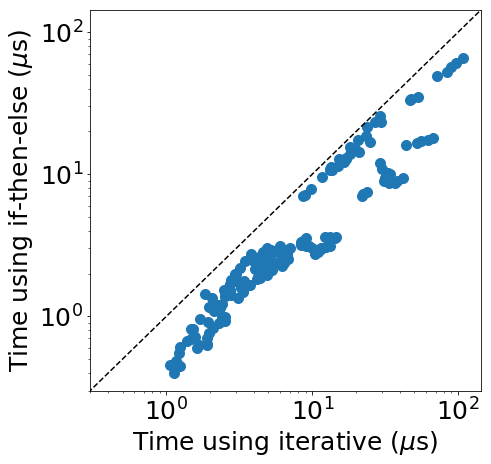}
    \vspace{-0.2cm}
    \caption{Time comparison for iterative and if-then-else decision trees.}
    \label{fig:time-iterative-ifelse}
\end{figure}

For decision trees, EmbML provides the options of using an iterative structure or if-then-else statements. In this analysis, the only difference between these options is structural and \change{does not} influence accuracy. As for memory usage, the amount of data to store is not affected, but code size may increase using if-then-else statements. However, both options achieved quite similar values for memory comparison: in the worst case, a classifier using if-then-else statements consumed only $2.55\: kB$ more memory than its iterative version -- a maximum increase of $6.04\%$. Therefore, we exclusively focus on comparing mean classification time, as displayed in Fig.~\ref{fig:time-iterative-ifelse}.

\change{Fig.~\ref{fig:time-iterative-ifelse} helps to recognize the lower time results of the if-then-else versions of the decision tree models. For this reason, we recommend choosing this version when generating a decision tree classifier with EmbML.}

\section{Comparison with related tools}
\label{sec:compare-related-tools}

To show that EmbML produces competitive classifiers, we evaluate its performance against classifiers generated by the following related tools: emlearn (version $0.10.1$), m2cgen (version $0.5.0$), sklearn-porter (version $0.7.4$) and weka-porter (version $0.1.0$). To make a consistent comparison, we selected only models from tools that have a direct correspondent in EmbML. For example, we consider the \textit{MLPClassifier} model since both EmbML and emlearn support it. The chosen models and tools that support them are listed below.

\begin{itemize}
    \item \textit{J48} is supported by EmbML and weka-porter;
    \item \textit{SVC} (polynomial kernel) is supported by EmbML, m2cgen, and sklearn-porter;
    \item \textit{SVC} (RBF kernel) is supported by EmbML, m2cgen, and sklearn-porter;
    \item \textit{LinearSVC} is supported by EmbML, m2cgen, and sklearn-porter;
    \item \textit{DecisionTreeClassifier} is supported by EmbML, emlearn, m2cgen, and sklearn-porter;
    \item \textit{MLPClassifier} is supported by EmbML and emlearn;
    \item \textit{LogisticRegression} is supported by EmbML and m2cgen.
\end{itemize}

\change{For decision trees, we employed their if-then-else versions provided by EmbML.} Also, we used the same trained model and test set to generate and evaluate the classifier versions. We executed each classifier with the datasets and microcontrollers to compare average classification time and memory usage. 

Our analysis incorporates only the time or memory values associated with a high accuracy to prevent including poor solutions -- \textit{e.g.}, the FXP16 versions of EmbML classifiers are faster but usually have lower accuracy than their corresponding FLT version. Therefore, after combining all outcomes from the same microcontroller, dataset, and classifier, we determined the average accuracy of these results, and eliminated those associated with an accuracy lower than it. 

With the remaining results from this process, we show in Table~\ref{tab:time-mem-related}, for each dataset, the number of cases that EmbML classifiers accomplished the best time and memory performances, and the total number of cases -- \textit{i.e.}, combinations of microcontrollers and classifiers -- that at least one classifier code was able to execute. Accordingly, the EmbML classifiers produced the best average classification time in at least $70.97 \%$ of the cases and the smallest memory consumption in at least $77.14 \%$. These results reveal that EmbML classifiers often perform better than the other solutions, indicating that EmbML is an advantageous alternative.

\begin{table}[htb]
    \centering
    \setlength{\extrarowheight}{-0.02cm}
    \setlength{\tabcolsep}{0.1cm}
    \caption{Overall time and memory comparison of classifiers from EmbML and related tools.}
    \label{tab:time-mem-related}
    \resizebox{0.99\columnwidth}{!}{%
    \begin{tabular}{cccc}
        \hline
        
        \hline
         \textbf{Dataset} & \makecell{\textbf{Cases which EmbML}\\ \textbf{classifiers achieve the}\\ \textbf{lowest time results}} & \makecell{\textbf{Cases which EmbML}\\ \textbf{classifiers achieve the}\\ \textbf{smallest memory results}} & \makecell{\textbf{Total number}\\ \textbf{of cases}} \\ \hline
         D1 & $25\:\: (71.43 \%)$ & $27\:\: (77.14 \%)$ & 35\\
         D2 & $27\:\: (75.00 \%)$ & $30\:\: (83.33 \%)$ & 36\\         
         D3 & $27\:\: (77.14 \%)$ & $30\:\: (85.71 \%)$ & 35\\
         D4 & $22\:\: (70.97 \%)$ & $27\:\: (87.10 \%)$ & 31\\
         D5 & $28\:\: (77.78 \%)$ & $35\:\: (97.22 \%)$ & 36\\
         D6 & $23\:\: (85.19 \%)$ & $21\:\: (77.78 \%)$ & 27\\ \hline
         \textbf{Total} & $152\:\: (76.00 \%)$ & $170\:\: (85.00 \%)$ & 200\\ \hline
         
         \hline
    \end{tabular}%
}
\end{table}

\section{Case study: Intelligent trap}
\label{sec:intelligent-trap}

\begin{figure}[htb]
     \centering
         \includegraphics[scale=0.25]{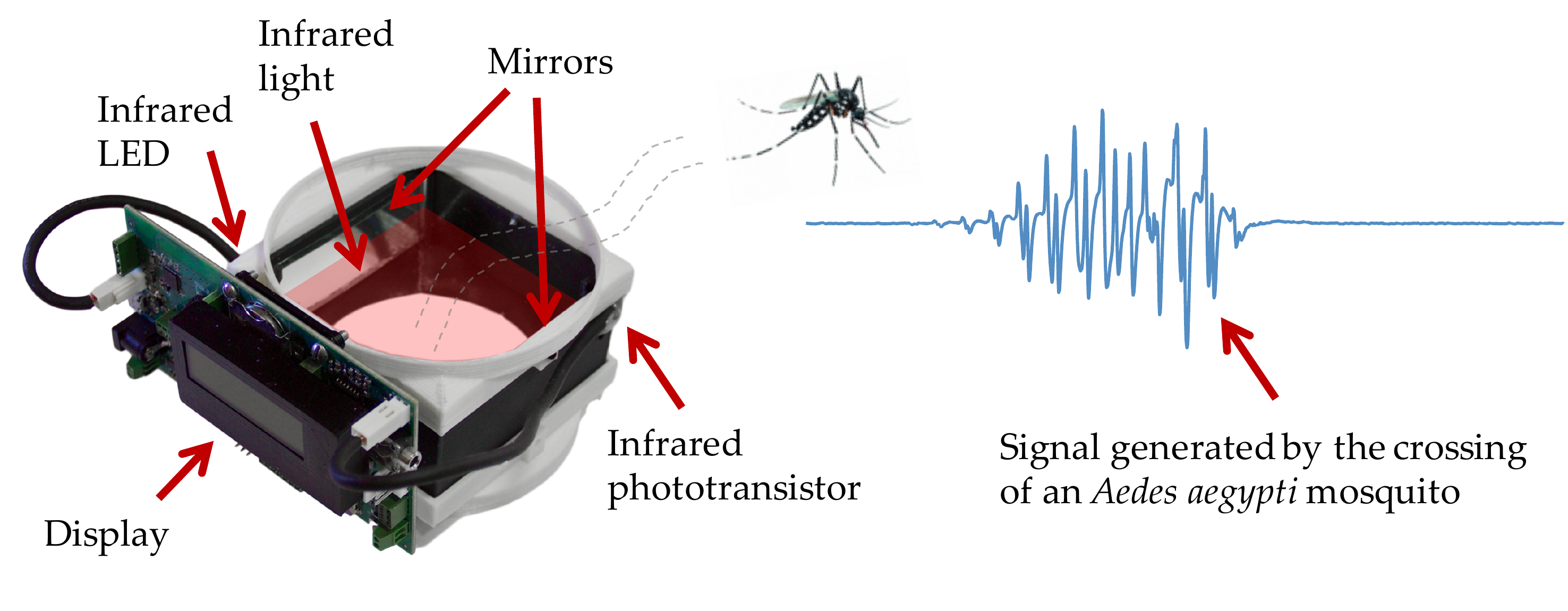}
     \vspace{-0.2cm}
     \caption{Intelligent trap sensor and a signal example.}
     \label{fig:sensor_signal}
 \end{figure}

In this section, we assess the EmbML classifiers using a practical application: an intelligent trap developed by our research group to classify and capture mosquitoes according to their species~\cite{batista2011sigkdd,de2013classification,chen2014flying,qi2015effective,silva2015exploring,SouzaChallenges:2020}. The main component of this trap is a low-cost optical sensor that remotely gathers data from flying insects. It consists of a structure of parallel mirrors, an infrared LED, and an infrared phototransistor~\cite{SouzaChallenges:2020}. When a flying insect crosses this structure, the movements of its wings partially occlude the light, producing small variations captured by the phototransistor as an input signal~\cite{batista2011sigkdd}. Fig.~\ref{fig:sensor_signal} show the sensor and a signal generated by an \textit{Aedes aegypti} mosquito. The trap's hardware processes the recorded signal to obtain features from its frequency spectrum, including frequency peaks, wingbeat frequency, and energy of harmonics~\cite{qi2015effective,silva2015exploring}. The classification process uses these predictive features to determine \change{the} species and gender of flying insects.

We used the Aedes aegypti-sex dataset~\cite{dos2018unsupervised} to train and evaluate the classifiers because its data was obtained with the same optical sensor of the trap. We performed a grid search to define the best set of hyperparameters for each classifier class supported by EmbML. After that, we used the test set to evaluate the outcomes and decided to select the \textit{J48} model with FXP32 format since it achieved the best results executing in the trap's microcontroller (MK20DX256VLH7): $98.92\%$ of accuracy, $1.26\: \mu s$ of average classification time, \change{$4.2\: kB$ of SRAM and $32.6\: kB$ of flash}. \change{On average, the trap consumes $435.6\:mW$ while waiting for insect crossings. When it processes and classifies an event, it consumes around $514.8\:mW$. Also, communication usually requires more $36\:mW$, using Bluetooth Low Energy (BLE).}

For each insect crossing event, we programmed the trap's microcontroller to perform the following tasks: read the input signal from the optical sensor; extract the predictive features from it; classify the event using the code produced by EmbML for the trained \textit{J48} model; and activate the trap's fan to capture female \textit{Aedes aegypti} and expel males.

\begin{figure}[htb]
    \centering
    \hfill
    \subfigure[Outside view of the cage]{\includegraphics[scale=0.32]{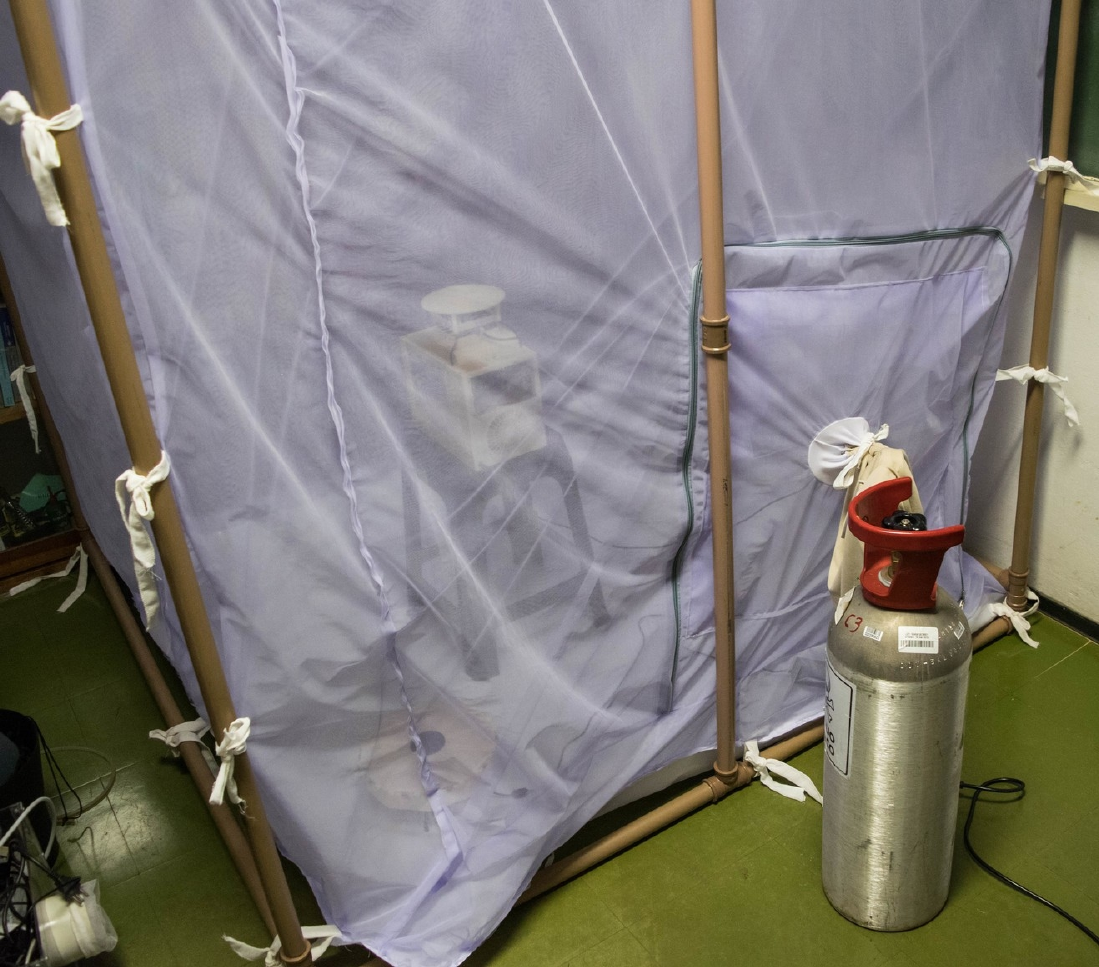}}
    \hfill
    \subfigure[Intelligent trap inside the cage]{\includegraphics[scale=0.455]{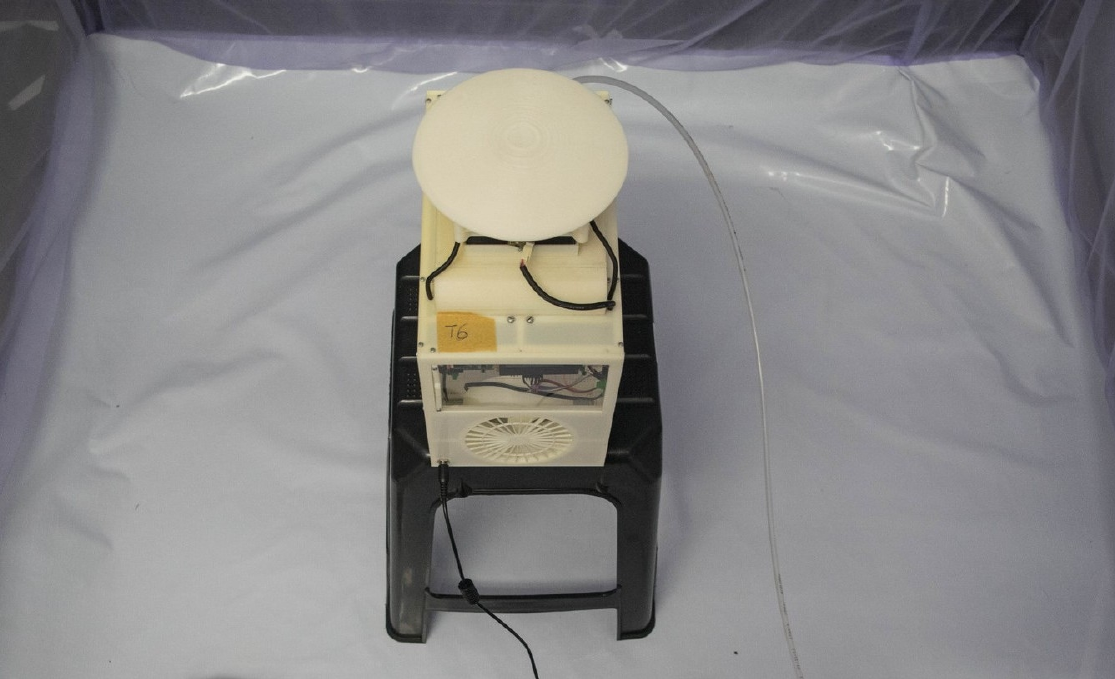}}
    \hfill
    \vspace{-0.2cm}
    \caption{Arrangement of the cage and the intelligent trap for mosquitoes employed in the experiments.}
    \label{fig:mosquitoes_cage}
\end{figure}

To analyze the trap performance \change{to capture mosquitoes selectively}, we used a cage that allowed releasing only \textit{Aedes aegypti} mosquitoes inside it, as illustrated in Fig.~\ref{fig:mosquitoes_cage}. It has dimensions of approximately $1.8\: m \times 1.8\: m \times 1.8\: m$ and includes double protection of mosquito netting fabric connected to a plastic pipe structure to create an isolated internal space. CO$_2$ was used to attract mosquitoes to the trap. We placed the cage in a room with controlled temperature and humidity. 

This experiment was conducted in three rounds of approximately 24 hours each. At the beginning of a round, we inserted 30 \textit{Aedes aegypti} mosquitoes (15 females and 15 males) inside the cage. At the end of every round, we manually captured, counted, and classified the insects outside and inside the trap, copied the produced event data, and reset the trap's microcontroller. Table~\ref{tab:trap-results} exhibits the results obtained at the end of each round of this experiment.

\begin{table}[htb]
    \centering
    \setlength{\tabcolsep}{0.15cm}
    \caption{Results from the intelligent trap experiment.}
    \label{tab:trap-results}
    \resizebox{0.99\columnwidth}{!}{%
    \begin{tabular}{cccccccc} \hline
    
        \hline
         \multirow{2}{*}{\textbf{Day}} & \multicolumn{2}{c}{\textbf{Inside}} & \multicolumn{2}{c}{\textbf{Outside}} & \multirow{2}{*}{\makecell{\textbf{Classified}\\\textbf{as Female}}} & \multirow{2}{*}{\makecell{\textbf{Total}\\\textbf{Captured}}} & \multirow{2}{*}{\makecell{\textbf{Total}\\\textbf{Events}}} \\ 
         
         & \multicolumn{1}{c}{\textbf{Female}} & \multicolumn{1}{c}{\textbf{Male}} & \multicolumn{1}{c}{\textbf{Female}} & \multicolumn{1}{c}{\textbf{Male}} & & &  \\ \hline
         
         \textbf{1} & 15 (100\%) & 3 (20\%) & 0 (0\%) & 12 (80\%) & 17 & 18 & 56 \\ 
         \textbf{2} & 15 (100\%) & 5 (33\%) & 0 (0\%) & 10 (67\%) & 17 & 20 & 34 \\ 
         \textbf{3} & 15 (100\%) & 7 (47\%) & 0 (0\%) & 8 (53\%) & 23 & 22 & 73 \\ \hline
    
    \hline     
    \end{tabular}%
    }
\end{table}

The results demonstrate that the trap was effective in capturing female mosquitoes. It caught all the released female mosquitoes, but it also wrongly captured at least $20\%$ of the males. This value is higher than expected from previous results, in which it was possible to classify correctly over $98\%$ of the examples. One explanation is that while capturing a female mosquito, \change{the trap could have caught some nearby males attracted by the female mosquitoes\cite{belton1979flight}.}

\section{Limitations and Future Work}
\label{sec:limitations}

\change{EmbML is an easy-to-use tool that automatically converts models into C++ code for microcontrollers. Such simplicity comes with a few limitations. One of them is to employ fixed values for $n$ and $m$ in FXP16 and FXP32 representations. Although the user can choose the values for these two parameters, they must remain constant during the entire classification process. This limitation is because EmbML, as a general and multi-model conversion tool, does not impose assumptions about the range of attributes values or require pre-processing steps such as data normalization.}

\change{In our experiments, we use the same values of $n$ and $m$ for all datasets. Such a decision simplifies the experiments but impacts the model accuracy since these values do not provide the best representation for each attribute. Alternatively, we could assume an initial representation for all model parameters and input values by enforcing data normalization and change the number of fractional bits as we operate over the model. Such an approach is prevalent in the literature (see, for instance, \cite{cerutti2020compact}).}

\change{Our literature review is limited to similar tools that support embedded classifiers. However, there exists a large body of literature that implements application-specific classifiers. In some cases, these classifiers may be more efficient than our solution because they are well-tuned for a given problem. However, the process of building such classifiers is usually laborious and may require in-depth technical knowledge.}

\change{Finally, EmbML does not implement approximation functions to RBF kernel as it does for the sigmoid function. In general, our experiments showed that models with such kernels present high demands for memory and processing. We will investigate forms of improving these classifiers. For instance, \cite{claesen2014fast} has proposed an approximation of the RBF function that can reduce the SVM RBF model complexity as well.}

\section{Conclusion}
\label{sec:conclusion}
We presented the implementation details of EmbML tool, which automatically produces classifier codes from trained models to execute in resource-constrained hardware. It includes specific support for this type of hardware: \change{avoidance of unnecessary use of SRAM, fixed-point operations for real numbers, approximations for the sigmoid function in MLP models, and if-then-else statements for representing decision trees.} Our experimental analysis empirically demonstrates that EmbML classifiers maintain the accuracy rates obtained with the training tools, improve performance by using available modifications, and achieve competitive results compared to related tools. Moreover, we successfully implemented an EmbML classifier in a practical application that allows \change{a trap to capture selectively disease-vector mosquitoes}.

\bibliographystyle{IEEEtran}
\bibliography{IEEEabrv,refs}

\begin{IEEEbiography}[{\includegraphics[width=1in,height=1.25in,clip,keepaspectratio]{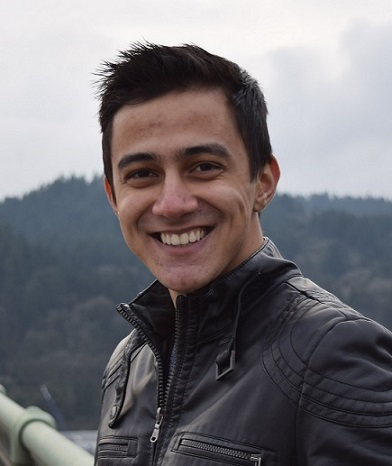}}]{Lucas Tsutsui da Silva} received the M.Sc. degree in Computer Science and Computational Mathematics (2020) from the University of S\~{a}o Paulo, Brazil and the B.Sc. degree in Computer Engineering (2016) from the Federal University of Mato Grosso do Sul, Brazil. His research interests include machine learning, data mining, and edge computing.
\end{IEEEbiography}
\vspace{-1cm}
\begin{IEEEbiography}[{\includegraphics[width=1in,height=1.25in,clip,keepaspectratio]{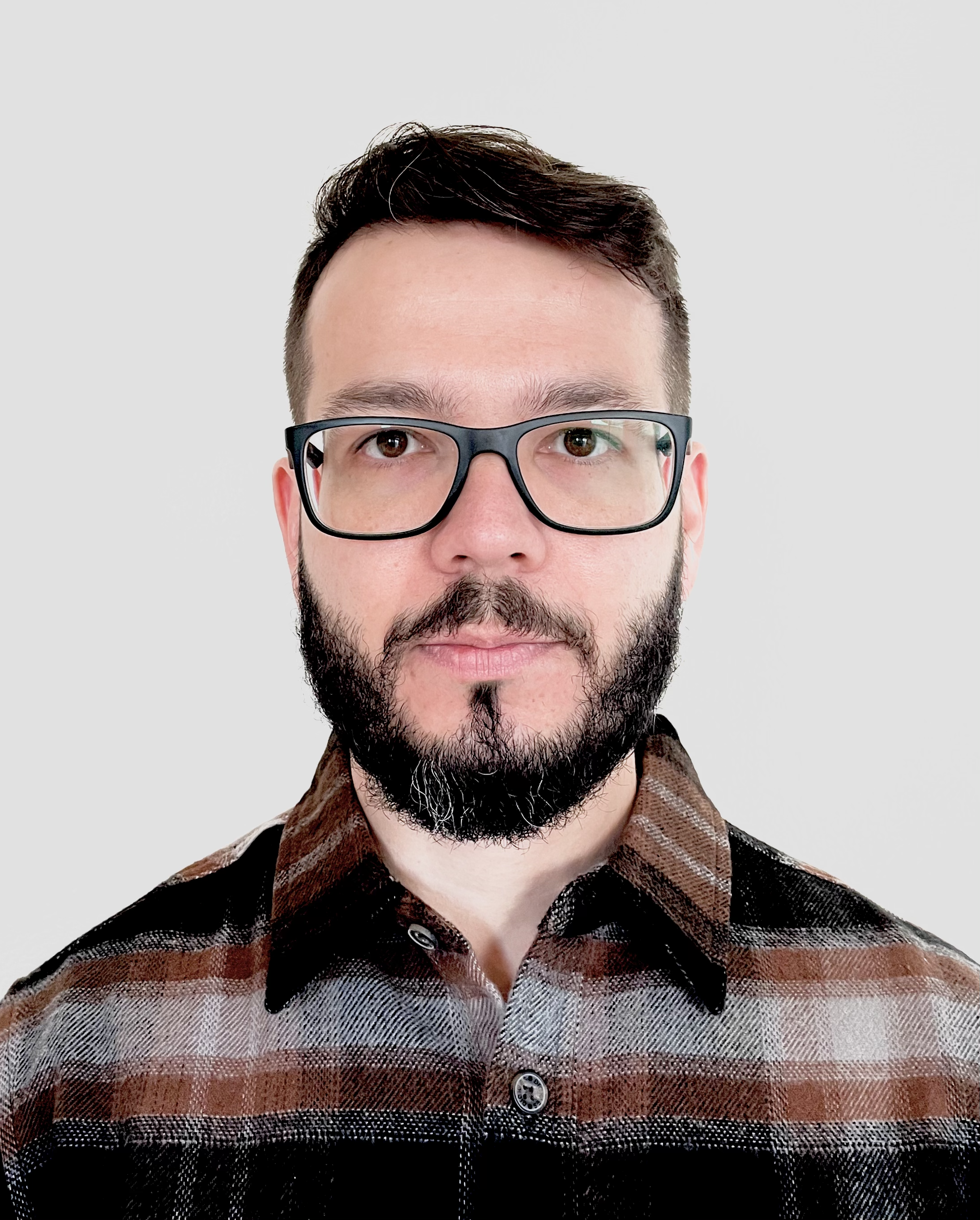}}]{Vinicius M. A. Souza} is an Associate Professor with the Graduate Program in Informatics (PPGIa) of the Pontifical Catholic University of Paran\'{a} (PUCPR), Brazil. Earlier, he was a Postdoctoral at the University of New Mexico, USA. He received the Ph.D. degree in Computer Science (2016) from the University of S\~{a}o Paulo, Brazil. Dr. Souza has authored over 40 articles in peer-reviewed conferences and journals, including Data Mining and Knowledge Discovery, Information Sciences, and IEEE-ICDM. His research interests include data mining, data streams, and time series.
\end{IEEEbiography}
\vspace{-1cm}
\begin{IEEEbiography}[{\includegraphics[width=1in,height=1.25in,clip,keepaspectratio]{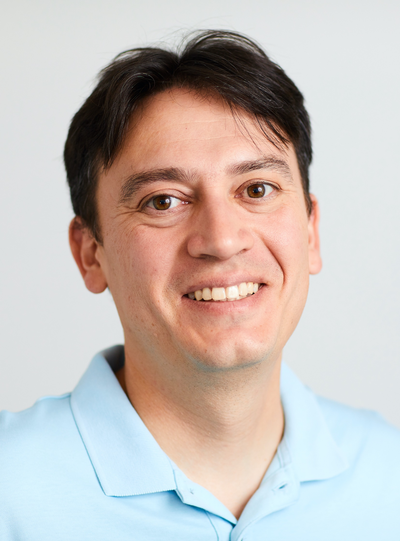}}]{Gustavo E. A. P. A. Batista} received the Ph.D. degree in Computer Science (2003) from the University of S\~{a}o Paulo, Brazil. In 2007, he joined the University of S\~{a}o Paulo as an Assistant Professor and became an Associate Professor in 2016. In 2018, he joined the University of New South Wales, Australia, as Associate Professor. Dr. Batista has more than 100 papers in conferences and journals. He has served as program committee member of conferences such as ACM-KDD, IEEE-ICDM and IJCAI, and as a member of the editorial board of the Machine Learning Journal. 
\end{IEEEbiography}

\end{document}